\documentclass[letterpaper]{article} % DO NOT CHANGE THIS
\usepackage{aaai24}  % DO NOT CHANGE THIS
\usepackage{times}  % DO NOT CHANGE THIS
\usepackage{helvet}  % DO NOT CHANGE THIS
\usepackage{courier}  % DO NOT CHANGE THIS
\usepackage[hyphens]{url}  % DO NOT CHANGE THIS
\usepackage{graphicx} % DO NOT CHANGE THIS
\usepackage{booktabs}
\usepackage{xcolor}
\usepackage{natbib}  % DO NOT CHANGE THIS AND DO NOT ADD ANY OPTIONS TO IT
\usepackage{caption} % DO NOT CHANGE THIS AND DO NOT ADD ANY OPTIONS TO IT
\usepackage{csquotes}
\usepackage{algorithm2e}
\RestyleAlgo{ruled}
\SetKwComment{Comment}{/* }{ */}

\def\ourmodel{GLaM }
\def\ourmodelnospace{GLaM}
\newcommand{\ourSystem}{\textsc{GLaM}}

\title{GLaM: Fine-Tuning Large Language Models for Domain Knowledge Graph Alignment via Neighborhood Partitioning and Generative Subgraph Encoding}

\author {
    Stefan Dernbach\equalcontrib\textsuperscript{\rm 1}, 
    Khushbu Agarwal\equalcontrib\textsuperscript{\rm 1},\\
    Alejandro Zuniga\textsuperscript{\rm 1}, 
    Michael Henry\textsuperscript{\rm 1}, 
    Sutanay Choudhury\textsuperscript{\rm 1}
}
\affiliations {
    \textsuperscript{\rm 1}Pacific Northwest National Lab\\
    902 Battelle Boulevard\\
    Richland, Washington 99354, USA
}

\begin{document}

\maketitle
\begin{abstract}
Integrating large language models (LLMs) with knowledge graphs derived from domain-specific data represents an important advancement towards more powerful and factual reasoning. As these models grow more capable, it is crucial to enable them to perform multi-step inferences over real-world knowledge graphs while minimizing hallucination.  While large language models excel at conversation and text generation, their ability to reason over domain-specialized graphs of interconnected entities remains limited. For example, can we query a LLM to identify the optimal contact in a professional network for a specific goal, based on relationships and attributes in a private database? The answer is no – such capabilities lie beyond current methods.  However, this question underscores a critical technical gap that must be addressed. Many high-value applications in areas such as science, security, and e-commerce rely on proprietary knowledge graphs encoding unique structures, relationships, and logical constraints. We introduce a fine-tuning framework for developing \textbf{G}raph-aligned \textbf{LA}nguage \textbf{M}odels ($\ourSystem$) that transforms a knowledge graph into an alternate text representation with labeled question-answer pairs.  We demonstrate that grounding the models in specific graph-based knowledge expands the models’ capacity for structure-based reasoning.  Our methodology leverages the large-language model's generative capabilities to create the dataset and proposes an efficient alternate to retrieval-augmented generation styled methods. 
\end{abstract}

\section{Introduction}
%Learning joint models of graph and language are useful for tasks that needs to combine the information from knowledge about a world or domain captured in natural language text with information that is observed in other structured forms of data.

%Graphs provide a way to query the entire knowledge that is not provided by a model which is trained to predict the most likely token. The logical structure provides an auditing mechanism.

Large language models (LLMs) have recently demonstrated disruptive potential with their ability to generate text and answer questions with human-like language proficiency. However, their reasoning remains limited by a reliance solely on textual training data, lacking integration with structured knowledge graphs encoding intricate real-world constraints and relationships. Bridging this divide by aligning LLMs with multi-relational graphs can enable grounded, factual inferences vital for applications driven by graph-structured data.

Past work on LLM-graph integration has predominantly focused on harnessing LLM knowledge to improve graph neural network performance on tasks like node classification and link prediction \cite{jin2023large}. The alternate direction of augmenting or ``fine-tuning" LLMs to reason over graphs has remained relatively unexplored. For instance, existing techniques still treat knowledge bases as external retrievable stores \cite{lewis2020retrieval}, rather than integrating them into model parameters.  Using the LLM as an encoder to transform text-based node and edge labels in a graph, and then fusing the LLM and GNN-derived representations has been the dominant approach for diverse applications ranging from product recommendation \cite{choudhary2022graph} to biomedical question-answering in a multiple-choice setting \cite{yasunaga2022deep}.

Our work is the first study on incorporating domain-specific knowledge graphs directly into LLM representations via fine-tuning, targeting accuracy improvements on open ended question answering(QA), a more complex task than the multiple choice setting explored in previous works. By encoding both schema and entities within specialized graphs like those in biomedical repositories, recommendation systems and social networks, we can enhance multi-hop reasoning grounded by real-world constraints. This addresses the challenge of factual hallucinations in free-form reasoning, while retaining versatile text handling strengths~\cite{touvron2023llama, nori2023capabilities}.

\begin{figure*}
    \centering
    \includegraphics[width=0.95\linewidth]{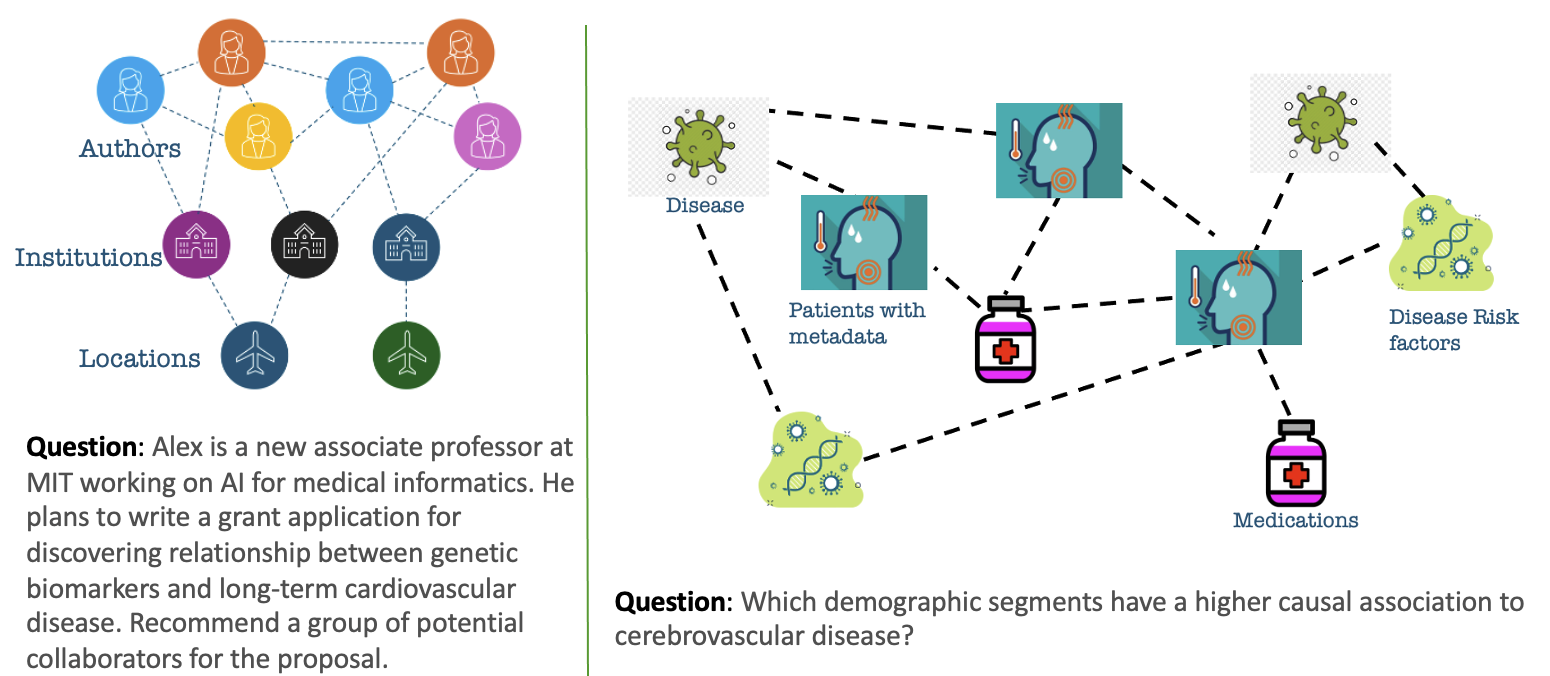}
    \caption{Motivating examples for aligning foundational models with domain-specific knowledge graphs.  The left figure demonstrates a query where a LLM needs to be integrated with a knowledge graph derived from a social network.  The right figure demonstrates a need where a LLM needs to be integrated with a patient-profiles to disease network extracted from an electronic healthcare records database.}
    \label{fig:motivation}
\end{figure*}

\subsection{Problem Definition}
 
Our work targets natural language question answering (QA) on graph data. We define $f_{LLM}: \mathcal{V} \rightarrow \mathcal{V}$ as a functional representation of a large language model that accepts a sequence of high-dimensional discrete tokens from a vocabulary $\mathcal{V}$ as input and produces an output sequence drawn from the same space. Given a natural language question $Q$ (also referred to as a prompt), $f_{LLM}(\cdot)$ tokenizes $Q$ into a sequence of tokens from $\mathcal{V}$ and returns an answer $A = f_{LLM}(\cdot)(Q)$.

Next, we introduce a graph dataset $G = (V, E)$, where $V$ is the set of vertices and $E$ is the set of edges. Importantly, we assume that $G$ was not included in the training data for $f_{LLM}(\cdot)$. Figure \ref{fig:motivation} describes real-world use cases that motivate these graph QA workloads, such as social or professional network-based recommendations or patient-specific clinical hypothesis generation.

% Next, we introduce a graph dataset $G = (V, E)$ where $V$ is the set of vertices and $E$ is the set of edges.  Most importantly, we assume that $G$ was not included in the training of $f_{LLM}(\cdot)$.  Figure \ref{fig:motivation} describes real-world drivers that motivates the formulation of such workloads.  In one of the applications, we seek to answer questions that require understanding of a social network structure to make effective recommendations.  The other query in Figure \ref{fig:motivation} exemplifies the need for clinical hypothesis generation that combines general purpose medical knowledge with information that changes from one patient network to another.

Our goal is to introduce a new function $f_{GLM}$ that utilizes information from $G$ to answer question $Q$. Formally, $A = f_{GLM}(G, Q)$. In this paper, we systematically explore following three query classes, in both open ended question answering and multiple choice setting:

\begin{enumerate}
    \item Fact recall: Evaluates GLM's ability to recall domain facts seen during training (e.g. answering "What are possible treatments for diabetes?" after seeing "Diabetes is treated with insulin and metformin").
    \item Inverse fact recall:  Assesses handling of relationship directionality, which recent work shows standard LMs struggle with ("A is B" does not imply "B is A") \cite{berglund2023reversal}. This is a key facet of graphs not previously explored for LLM-graph models. 
    %\item Relationship inference: Requires bridging non-direct entity connections, and multi-hop reasoning (vs. single-hop fact recall).
    \item Chain-of-Reasoning: Complex queries such as Figure \ref{fig:motivation} (left) that necessitate appropriately employing graph structure knowledge.
\end{enumerate}

\subsection{Technical Approach and Related Work}

Exploring the intersection of large language models and knowledge graphs has strong significant interest over the past few years. We begin by outlining key design paradigms from literature for answering complex reasoning queries on knowledge graphs by aligning graph data with large language models (LLMs) and refer the reader to a collection of excellent survey articles for a detailed overview of this emerging sub-field \cite{pan2023unifying, liu2023towards, jin2023large}. Any approach must address two questions: 1) how to encode graph $G$ into the LLM's knowledge representation, and 2) how query $Q$ is executed. 

\textbf{Delegation to a GNN} A common approach uses a graph neural network as encoder.  Given a natural language query $Q$ this requires extracting entities and relations from the query and integrate GNN and LLM representations.  Such integration can be done via learning a joint-model coupling the LLM and GNN representations \cite{saxena2020improving, yasunaga2022deep} or using a soft prompting approach that inserts a GNN-derived vector embedding into the LLM prompt \cite{tian2023graph}. 

\textbf{Retrieval Augmented Generation} Retrieval augmented generation (RAG)  approaches follow a similar path of implementation.  The difference here is that instead of delegating to a GNN, an external graph database \cite{jiang2023structgpt} or a vector database \cite{tian2023graph} containing node and/or relation embeddings is queried .  In both approaches, the LLM is used as an routing interface to a native graph database or machine-learning model, and the answers from the appropriate graph-based component is fed back to user with LLM serving as generative layer that produces the final answer.  

\textbf{Few-Shot Prompting} In this approach subgraphs relevant to $Q$ are extracted and inserted into the prompt with examples \cite{fatemi2023talk}. While promising, this approach faces potential drawbacks requiring encoding full graphs in LLM prompts or performing multi-hop subgraph retrievals for each question.

\textbf{Motivation for Fine-Tuning} Irrespective of their differences, all of the above approaches potentially face a fundamental limitation - they cannot contextually integrate symbolic constraints to shape intermediate reasoning. Approaches retrieving embeddings or triples solely based on the initial query overlook multifaceted execution where dynamic routing with mixture-of-experts \cite{shazeer2017outrageously, zhou2022mixture}, planning \cite{hao2023reasoning, yao2023tree} and heuristics search \cite{sprueill2023monte, sun2023think} steps modify information needs in every reasoning step. Fixed retrieval precludes dynamically folding graph structure into each decision point.

In contrast, fine-tuning instills domain knowledge into model parameters and representations a priori, rather than treating the graph as an external add-on. By encoding constraints and dependencies directly into the knowledge substrate, fine-tuning allows contextual graph influence at each step of modeled cognition. Rather than acting as a static look-up, the graph becomes an integral inference component - shaping complex reasoning in a tighter, more fine-grained fashion.

\subsubsection{Our Approach and Contributions}
We introduce an algorithm to iteratively partition and encode the neighborhood subgraph around each node into textual sentences for fine-tuning data. This transforms graph structure into a format that large language models can ingest and fine-tune. We explore encoding strategies on two graphs: 1) UMLS - a biomedical knowledge base, and 2) DBLP - an academic publication network.

Our work makes the following contributions.  
\begin{enumerate}
    \item  Our neighborhood partitioning and encoding scheme accommodate real-world graph properties like skewed size distributions and sparsity. Our approach opens up future experimental possibilities where encoding is tuned for LLMs by setting context size limits based on cost-accuracy tradeoffs.
    \item We propose and assess five encoding approaches leveraging the LLM's innate summarization and text generation strengths. For example, we evaluate neighborhood summaries produced by the LLM. Encouragingly, our results align with similar methods from concurrent work \cite{fatemi2023talk}, confirming the promise of this direction.
    \item We developed a new domain question answering dataset based on two above graphs with a suite of evaluation tasks capturing link prediction to multi-hop reasoning queries. The code and datasets will be released as open source upon acceptance. %comprising \textbf{X} queries.  The code and datasets will be released as open source upon acceptance.  
\end{enumerate}

\section{Methods}
\label{sec:methods}
\textbf{Task Definition} We propose methods for transforming a knowledge graph into a corresponding text-based fine-tuning dataset for language models. Our goal is to produce pairs of (context, question-answer) \cite{ouyang2022training, wei2021finetuned} that require neural-graph reasoning to answer open domain questions involving relational or multi-hop reasoning.  

% Each generated question 
 
% Specifically, we focus on generating questions that evaluates the resulting fine-tuned language model's ability to retrieve/remember domain facts, infer symmetrical relationships,and infer new facts that require multi-hop reasoning. 
% \begin{enumerate}
%     \item Relational/multi-hop question: entity [relation] ?
%     \item Multiple choice: entity [relation] ? : List[options]
% \end{enumerate}

We begin with describing a generic algorithm (Algorithm \ref{alg:gen_finetuning_dataset}) that encodes a node's k-hop neighborhood into such a context and QA-pair through a composition of multiple operator functions.  We discuss the implementation of these operators in finer detail in the later half of the section.

\begin{algorithm}
	\SetAlgoLined
	\SetKw{Initialize}{Initialize}
\textbf{Require:} Graph $G$ with nodes set $V$ and edges set $E$,
context subgraph node limit $N_{max}$

    Fine-tuning dataset $D\leftarrow \emptyset$\\
	\For{\textbf{each} $v \in V(G)$}{
            $G_{context}(v, k) = f_{aggr}(G, v, k)$
            
            $partitions=f_{partition}(G_{context}(v, k), N_{max}$) 
            
            \For{\textbf{each} $g_{sub} \in partitions$}{
            
                $X_{context} = f_{enc}(g_{sub})$
                
                $X_{qa} = f_{qa}(g_{sub})$
                
                $append(D, concat([X_{context}, X_{qa}])$
            }
        }
        return $D$
	\caption{Fine-tuning dataset generation.}
	\label{alg:gen_finetuning_dataset}
\end{algorithm}

\subsection{Optimal Generation of Subgraph Contexts}

For every node $v \in V(G)$, we transform the $k$-hop neighborhood of $v$ into a set of pairs of the form: $(f_{enc}(G, v), f_{qa}(G, v))$. Algorithm \ref{alg:gen_finetuning_dataset} describes a step-by-step in which we iterate over every node in the graph and encode it's k-hop neighborhood subgraph, denoted as $G_{context}(v, k)$ into the alternate text-based representation.  

\begin{enumerate}
    \item We retrieve the k-hop neighborhood subgraph as $G_{context}(v, k)$ using a query function denoted as $f_{aggr}(\cdot)$. 
    \item $f_{enc}$ encodes $G_{context}(v, k)$ or its partitioned subgraph into text. 
    \item $f_{qa}(G, v)$ generates QA pairs requiring reasoning on $G_{context}(v, k)$. Same subgraph is used to drive the inputs for $f_{enc}(G, v)$ and $f_{qa}(G, v)$.
    \item The concatenated output of $f_{enc}(G, v)$ and $f_{qa}(G, v)$ is a text sequence of discrete tokens $X_v$ drawn from $\mathcal{V}$, the vocabulary of the LLM function $f_{LLM}(\cdot)$ mentioned previously.
    \item Any LLM function $f_{LLM}(\cdot)$ needs to operate within a maximum token limit constraint (denoted as $T_{max}$).  We partition $G_{context}(v, k)$ to respect LLM token limits $T_{max}$ such that $len(X_v) < T_{max}$. 
\end{enumerate}

We introduce a hyperparameter $N_{max}$ to partition $G_{context}(v, k)$ into subgraphs within node count $N_{max}$. This prevents tokenized sequence lengths from exceeding $T_{max}$.  Choosing an optimal $N_{max}$ is key because degree distributions in $G_{context}(v, k)$ can be highly skewed. Given cost constraints associated with $T_{max}$, we want to pick $N_{max}$ and encoding strategies that maximize context lengths for the LLM's capabilities.

% We introduce a parameter $N_{max}$ such that $G_{context}(v, k)$ is partitioned into multiple subgraphs via a function $f_{partition}(\cdot)$, and the node count of each partitioned subgraph do not exceed $N_{max}$. $N_{max}$ can be chosen empirically or learnt through hyperparameter tuning such that the tokenized representation of each partition and a corresponding question-answer pair never exceeds $T_{max}$, the maximum token limit for the LLM.  

% The optimal selection of $N_{max}$ is an important parameter considering that the distribution of node counts in $G_{context}(v, k)$ can be highly skewed in a graph with high degree nodes. Given that the choice of $T_{max}$ is typically driven by cost considerations, we will want to select a $N_{max}$ and an encoding strategy to generate $G_{context}(v, k)$ that maximally leverages the LLM's capability to operate on large context sizes.

\subsection{Neighborhood Encoding Functions}
\label{sec:graph2text}

\begin{figure*}[h]
  \centering
  \includegraphics[width=0.8\textwidth]
  {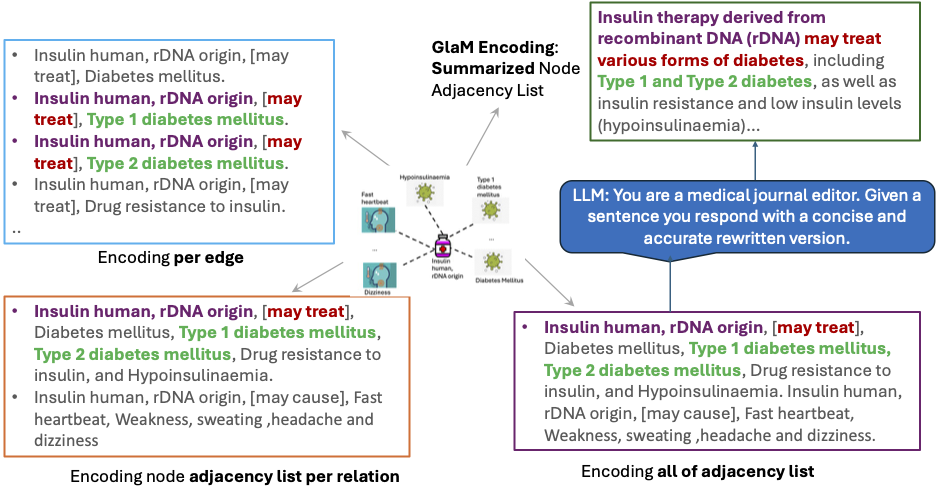}
\caption{Illustration of Graph Encodings in GLaM: Top left box shows "Encoding via triples", where each line represents an edge mapped to one training sample. The bottom left box shows graph encoding when given a node and relation, all relevant entities are collated into single training sample. The bottom right box shows when all relations/edges corresponding to a node are coalesced into single training sample. and top right box demonstrates the impact of summarization on the training sample.  Summarizing helps to 1) map unwieldy node labels into human interpretable form, 2) reduce redundant terms, and 3) reduce overfitting to frequent node and edge labels. Collectively this leads to better semantic alignment betweeen the knowledge graph and LLM's vocabulary and improves resulting model performance in all graph tasks.}
\label{fig:graph_encoding}
\end{figure*}

The purpose of a neighborhood encoding function is to translate the neighborhood subgraph $G_{context}(v, k)$ centered around a node $v$ into a textual representation that can be effectively processed by a large language model (LLM). This process is crucial for enabling the LLM to perform higher-order reasoning and answer complex questions about the graph.

There are two main factors that influence the choice of a neighborhood encoding function:
\begin{enumerate}
    \item \textbf{Communicating Graph Structure and Higher-Order Reasoning Requirements to the LLM}: The encoding function should effectively capture the structural relationships between nodes in the subgraph, as well as any higher-order logical dependencies that may exist. This can be achieved by incorporating information about the edges and their types, as well as the relationships between multiple nodes.
    \item \textbf{Semantic Alignment with the LLM's Internal Knowledge Representation}: The encoding should represent the nodes and relations in the graph in a way that is consistent with how the LLM stores and interprets information. This can involve using natural language labels for nodes and edges, or generating descriptive labels using a node's neighborhood when node labels are not recognizable to a LLM (such as an academic network), while ensuring that the encoded representation preserves the semantic meaning of the graph elements.
\end{enumerate}
  
\textbf{Encoding via Triples}: A simple approach to neighborhood encoding is to translate the edge data into (source, relation, target) triples. This provides the LLM with basic information about the relationships between nodes, but it is limited to representing only single edges per training sample and has limited context size.
%This limits during training types of questions that can be answered to fact recall and link prediction.

% \textbf{Encoding via Relational Groups} Another approach is to collate all target nodes corresponding to a source node and relation pair. 

\textbf{Encoding via Adjacency List/Relational Groups} To enable the LLM to perform more complex reasoning tasks, we update the neighborhood encoding to include information about multiple nodes in the subgraph. We experiment with two different options: including the entire adjacency list of the central node $v$, and by partitioning the neighbors into subsets based on their relation types. We observe that more sophisticated approaches, such as sampling techniques are relevant for large neighbor lists but are not implemented in current work.

\textbf{Encoding via Summarization} Next, we focus on the semantic alignment objective and use prompting methods to rewrite the encoding from above methods into more coherent representations (Figure \ref{fig:graph_encoding}). 
% Figure ~\ref{fig:graph_encoding} demonstrates the impact of using summarization prompt on a sample node context generated during training of a medical knowledge graph. 
\begin{itemize}
    \item The promptimg allows us to map unwieldy node labels to human understandable terms: For example,``Insulin human, rDNA origin” is mapped by LLM to ``Insulin therapy from recombinant DNA” allowing for better interpretation during fine-tuning.
    \item It reduces redundant text from similarly labeled nodes: ``Diabetes mellitus, Type 1 diabetes mellitus, Type 2 diabetes mellitus" is mapped to ``diabetes, including Type 1 and Type 2 diabetes."
    \item Introduces additional knowledge/synonyms into training: ``Hypoinsulinaemia” is mapped to ``low insulin levels (hypoinsulinaemia)," and ``rDNA” is expanded to ``recombinant DNA."
    \item Prompt-based rewriting also helps reduces to address overfitting training to only a few relation labels, by mapping them to different phrases.  Examples of such overfitting was observed with the ``may treat" relationship, where the high number of occurrence of this phrase in a specific pattern causes the LLM to generate answers incorrectly filled with too many occurrences of the ``may treat" phrase.

\end{itemize}

\textbf{Encoding via Node Descriptors}  The previous encoding step leveraged the LLM's understanding of specific entities (such as ``rDNA") to rewrite with maximal semantic alignment.  However, training on new graph data can include unfamiliar terms to the LLM, i.e. words or phrases that appear rarely or do not occur at all in initial training. A common example of this problem involves encoding the names of people not common to standard LLM training datasets. Also, we do not want to map a person based on their name, but account for their profile attributes or k-hop connectivity in the network.  We generalize this need by transforming the k-hop context subgraph ($G_{context}(v, k)$) into a set of text-based node descriptors by leveraging the LLM's zero-shot capabilities.  Typically, this is a step where an alternate implementation would have retrieved a GNN representation. For example, to expand on the information about authors in the DBLP dataset, we prompt the LLM to extract the topic areas of paper abstracts and construct a list of topics the author has published on from their paper history.

\textbf{Generating Question-Answer Pairs} Finally, given a text context generated from a subgraph $G_{context}(v, k)$, we generate a set of question-answer pairs via prompting the text context for different tasks (fact recall, inverse fact recall, multi-hop question answering).  Each of the questions are also mapped into two style of answers: 1) open-domain question-answering, and 2) multiple-choice questions.  For example, given a (head, relation, tail) triple as the subgraph context, its multiple choice answer candidates are generated by including one of the tail entities and a random selection of other nodes in the graph to form a set of possible answers to the question.

\section{Experiments}

\begin{table*}[!h]
\centering
\begin{tabular}{cccccccccc}
     & \multicolumn{3}{c}{Fact Recall} & \multicolumn{3}{c}{Reverse Recall} & \multicolumn{3}{c}{Multi-hop Reasoning} \\
     & P & R & F1 & P & R & F1 & P & R & F1 \\
     \toprule
     Llama 7B Chat & $0.594$ & $0.631$ & $0.608$ & $0.382$ & $0.519$ & $0.439$ & $0.595$ & $0.631$ & $0.609$\\
     \ourmodel(Triples) & $0.683$ & $0.597$ & $0.627$ & $\mathbf{0.431}$ & $0.533$ & $\mathbf{0.474}$ & $0.677$ & $0.589$ & $0.621$ \\
     \ourmodel(Relational Grouping) & $0.679$ & $0.678$ & $0.673$ & $0.403$ & $\mathbf{0.537}$ & $0.459$ & $0.662$ & $0.663$ & $0.657$ \\
     \ourmodel(LLM Summarization) & $\mathbf{0.724}$ & $\mathbf{0.725}$ & $\mathbf{0.720}$ & $0.386$ & $0.527$ & $0.445$ & $\mathbf{0.689}$ & $\mathbf{0.696}$ & $\mathbf{0.688}$ \\
     \midrule
     Llama 13B Chat & $0.699$ & $0.623$ & $0.652$ & $0.396$ & $0.529$ & $0.451$ & $0.695$ & $0.623$ & $0.650$ \\
     \ourmodel 13B (LLM Summarization) & $0.708$ & $0.730$ & $0.714$ & $0.395$ & $0.534$ & $0.453$ & $0.675$ & $0.697$ & $0.681$ \\
    \bottomrule
\end{tabular}
\caption{\textbf{UMLS Results} comparing the baseline Llama LLM with three versions of a refined \ourmodel on questions generated from the UMLS knowledge graph. Each version corresponds to an encoding strategy described in the Methods section. Precision (P), Recall (R), and F1 scores are reported using Bert scores.}
\label{tab:umls}
\end{table*}

\begin{table*}[!h]
\centering
\begin{tabular}{cccccccccc}
     & \multicolumn{3}{c}{Fact Recall} & \multicolumn{3}{c}{Reverse Recall} & \multicolumn{3}{c}{Multi-hop Reasoning} \\
     & P & R & F1 & P & R & F1 & P & R & F1 \\
     \toprule
     Llama 7B Chat & $0.174$ & $0.177$ & $0.175$ & $0.168$ & $0.173$ & $0.170$ & $0.168$ & $0.171$ & $0.169$\\
     \ourmodel (Triples) & $0.105$ & $0.103$ & $0.104$ & $0.103$ & $0.102$ & $0.102$ & $0.100$ & $0.099$ & $0.099$\\
     \ourmodel (Relational Grouping) & $0.259$ & $0.261$ & $0.259$ & $0.259$ & $0.264$ & $0.260$ & $0.256$ & $0.259$ & $0.257$\\
     \ourmodel (Adjacency List) & $0.255$ & $0.258$ & $0.255$ & $0.247$ & $0.252$ & $0.249$ & $0.251$ & $0.253$ & $0.251$\\
     \ourmodel (Node Descriptors) & $0.313$ & $0.312$ & $0.312$ & $0.309$ & $0.314$ & $0.311$ & $0.318$ & $0.316$ & $0.316$\\
     \ourmodel-7B & $\mathbf{0.424}$ & $\mathbf{0.426}$ & $\mathbf{0.424}$ & $\mathbf{0.401}$ & $\mathbf{0.407}$ & $\mathbf{0.402}$ & $\mathbf{0.409}$ & $\mathbf{0.410}$ & $\mathbf{0.408}$\\
     \midrule 
     Llama 13B Chat & $0.150$ & $0.155$ & $0.152$ & $0.144$ & $0.151$ & $0.147$ & $0.153$ & $0.159$ & $0.155$ \\
     \ourmodel-13B  & $0.446$ & $0.446$ & $0.445$ & $0.381$ & $0.385$ & $0.382$ & $0.398$ & $0.398$ & $0.397$ \\
    \bottomrule
\end{tabular}
\caption{\textbf{DBLP Results} comparing the baseline Llama LLM with five versions of  \ourmodel on questions generated from the DBLP citation graph. \ourmodelnospace-7B/13B represents a combination of  strategies: aggregation of node descriptors, utilizing adjacency lists as context and performing summarization.  Precision (P), Recall (R), and F1 scores are reported using Bert scores.}
\label{tab:dblp}
\end{table*}

\begin{table*}[!h]
\centering
\begin{tabular}{ccccccc}
    & \multicolumn{3}{c}{UMLS} & \multicolumn{3}{c}{DBLP} \\
    & Fact & Reverse  & Multi-hop & Fact & Reverse & Multi-hop  \\
    & Recall & Fact Recall & Reasoning & Recall & Fact Recall &  Reasoning \\
    \toprule
    Llama 7B Chat & $61.23$ & $57.21$ & $61.1$ & $35.26$ & $36.19$ & $27.99$\\
    \ourmodel-7B-MC & $100$ & $59.71$ & $91.93$ & $78.62$ & $75.68$ & $73.34$\\
    \bottomrule
\end{tabular}
\caption{\textbf{Multiple Choice Results} Comparison of LLM and \ourmodel accuracy for fact recall, reverse fact recall, and fact inference on the UMLS and DBLP graphs.}
\label{tab:mc}
\end{table*}

In this section, we address following research questions (RQ) through experimental analysis:
\begin{enumerate}
    \item \textbf{RQ-1} Does finetuning using graph encoding improve an LLM's ability to recall the facts?
    \item \textbf{RQ-2} Does finetuning an LLM with graph encoding improve its ability to answer open-domain natural language questions through performing multi-hop reasoning on the graph domain?
    \item \textbf{RQ-3} Which strategies for encoding the subgraph context yields maximal semantic alignment of the original LLM and the target graph?
    % \item \textbf{RQ-4} How do the results of the above questions translate to multiple-choice questions?
\end{enumerate}

%\subsection{Experimental settings}
\subsection{Datasets}

We present the results of training \ourmodelnospace s on two graph datasets, DBLP~\cite{Tang:08KDD} and UMLS~\cite{bodenreider2004unified}, with diverse applications and coverage in LLMs to demonstrate the response improvement over the baseline language models. 

\textbf{Unified Medical Language System\footnote{https://www.nlm.nih.gov/research/umls} (UMLS)}~\cite{bodenreider2004unified} is a medical knowledge graph. We use a processed version of the knowledge graph from Yasunaga et al.~\cite{yasunaga2022deep} consisting of $297,927$ concepts, $98$ relation types and $1,212,586$ edges that capture relationships across a breadth of medical concepts. For \ourmodel training, we select a subgraph that captures relationships between different diseases, symptoms and medications. This results in a reduction to 4 different relation types: ``\textit{cause of}", ``\textit{may cause}", ``\textit{risk factor of}", and ``\textit{may treat}" totalling $126,149$ triples. 

\textbf{DBLP\footnote{https://www.aminer.org/citation}}~\cite{Tang:08KDD} is a citation graph dataset extracted from DBLP, ACM, MAG, and other sources. The dataset includes paper citations, abstracts, authors, publication years, venues, and titles.  For training the \ourmodel we focus on the set papers containing titles, abstracts, venues, and $2$ or more authors, leading to  $19,577$ unique papers. 

\subsection{Training and Inference setup}

For both UMLS and DBLP, the extracted natural language questions and answers were split into 70\% training (fact recall) and 30\% test (multi-hop reasoning). We used Microsoft Deepspeed framework ~\cite{rasley2020deepspeed} for supervised prompt and response fine-tuning. A grid-search was performed over training hyper-parameters using Llama-7b-chat-hf as the base model for training \ourmodelnospace. A learning rate of $1e-5$ and a cosine learning rate scheduler were used with the fused Adam optimizer with bfloat16 precision. The maximum sequence length was set to $256$ and a maximum of $4$ training epochs were used for all models. A cluster of 8 A100 GPUs with 80GB of GPU memory each were used for training with a per-device batch size of 4 questions resulting in a total training batch size of 32. We use Llama-2-7b-chat-hf and Llama-2-13b-chat-hf~\cite{touvron2023llama} models from Hugging Face as the baseline models for training. Training the 7b model on UMLS takes approximately 9 minutes and 16 minutes for the 13b model. For DBLP, training time is approximately 11 and 21 minutes respectively.

\subsection{Evaluation Tasks}
\textbf{Fact recall}: This task is equivalent to question answering tasks in language models and test \ourmodelnospace's ability to remember domain level facts seen during training. For example, given a training sentence such as `\textit{`Diabetes is treated with insulin and metformin}" (from UMLS), the model is queried for ``\textit{What are possible treatment of diabetes?}". Similiarly, for the DBLP dataset given a sentence such as ``\textit{[Students learn CS in different ways: insights from an empirical study] was written by Anders Berglund.}", the model is queried with ``\textit{[Students learn CS in different ways: insights from an empirical study] was written by whom?}" The UMLS question set for fact recall contains $7710$ questions and the DBLP set contains $13,704$.

\textbf{Inverse Fact Recall}: This task is equivalent to reverse question answering tasks~\cite{berglund2023reversal} in language models and test \ourmodel's ability to infer reverse relationships from the domain level facts seen during training. For example, given the above training statement, the model is queries for ``\textit{Which disease can be treated with insulin?}" There are $11130$ questions in the UMLS reverse fact recall question set and $13704$ in the DBLP set.

\textbf{Multi-hop Reasoning}:  This task mirrors the link prediction task in a GNN setting and tests the \ourmodelnospace 's ability to infer new facts (graph edges) by reasoning over facts seen during training. The UMLS question set for multi-hop reasoning contains $3347$ questions and the DBLP set contains $5873$.  A common style of question we explore for DBLP is that of recommending authors to collaborate with.  Using the DBLP question referred to in the fact recall task as example, a multi-hop reasoning question would ask: ``\textit{Anders Berglund would like to write a paper titled [Students learn CS in different ways: insights from an empirical study] to publish in Proceedings of Australasian computing education. Who should they work with and why?}"

\textbf{Multiple-choice}: Each evaluation task: fact recall, inverse fact recall, and multi-hop reasoning, are reformatted as multiple choice questions.   Question includes the correct answer and four additional incorrect options randomly selected from the graphs respectively. Note that this is a much easier task than open ended question answering setting, requiring models to only pick the most likely answer out of given options.

\subsection{Evaluation Metrics} 

To account for the inherent text variability in LLM or \ourmodel generated answers, we use the BERTScore metric \cite{zhang2019bertscore} for open-ended domain QA setting and accuracy for multiple choice questions. %In addition, we also perform a graph metrics based evaluation to compare the distance between ground truth answers to LLM and \ourmodel generated answers during link prediction. 

\textbf{Bert Score:} Compares text similarity between the model generated response to the expected response. The microsoft/deberta-xlarge-mnli model \cite{he2020deberta} is used for calculating BertScore for it's strong performance in natural language understanding (NLU) tasks. We report precision (P), recall (R), and F1 scores across the evaluation set.

\textbf{Accuracy:} We use the standard accuracy measure to evaluate a model's ability to identify the correct answer out of 5 possible choices in a multiple choice setting.
%STEFAN: We have nothing on shortest path distance
%\textbf{Shortest Path Distance (SPD):} Measure the length of the given answer from the original author in the graph. %The correct answer has a graph distance of $0$ while an incrorrect answer has an integer SPD $>=1$. An answer not in the graph is \SD{TBD!}
%\paragraph{Qualitative Evaluation} is provided in the appendices for each model and dataset. 
%\paragraph{Training Time etc.}

\subsection{Results}

Results for training \ourmodel are presented in Table~\ref{tab:umls} and Table~\ref{tab:dblp}. We discuss the results on the individual datasets and then provide overall conclusions.

\textbf{UMLS} graph experiment results are given in Table~\ref{tab:umls}. %Fact recall evaluates the LLM/\ourmodel using the same questions as used for training the \ourmodel. These test cases demonstrates the \ourmodel's ability to retain information. Reverse facts uses the same set of facts as the recall questions but the questions are reversed, i.e. instead of asking 'X may cause what?' with answer 'Y', the question asks 'Y may be cause by what?' with answer 'X'. Finally, the fact inference questions are additional facts from the graph that were withheld during training in order to evaluate the \ourmodel's inference capabilities. We use Llama-7b-chat-hf as our baseline LLM and evaluate against the 3 \ourmodel training paradigms outlined above trained from this model. We also compare Llama-13b-chat-hf and a \ourmodel trained with using it as a baseline.
For both fact recall and inference, using LLM based summarization encoding to rewrite the statements exhibits the best performance across precision, recall and F1 scores. However, for reverse fact recall, using the simpler training approaches leads to a slight improvement in scores. All fine-tuned \ourmodel versions outperform the baseline LLM showing that even naive training strategies offer some improvement over the baseline LLM. While the 13b version of Llama outperforms its 7b counterpart, once trained, there is negligible difference between the 13b and 7b \ourmodel.

\textbf{DBLP} citation graph experimental results are given in Table~\ref{tab:dblp}. The  \ourmodel version with complete adjacency and LLM based summarization achieves the best results across all tasks. Unsurprisingly, the untrained LLM did only moderately better than random guessing for the multiple choice task because of the number of unfamiliar names in the dataset. There is also a general trend of improved performance as neighborhood information is collated into the training, with the exception of adding the venue of the publication not having a noticeable affect,likely due to  title being sufficient to capture a publications context. There is a slight improvement of the 13b version of \ourmodel over the 7b version for fact recall but the 7b version slightly outperforms the larger \ourmodel on the reverse fact recall and fact inference tasks. This combined with similar findings on UMLS indicate that the smaller LLM is sufficient for fact retention and inference when fine-tuned for the domain.

\textbf{Multiple Choice} results for both UMLS and DBLP are provided in Table~\ref{tab:mc}. Across all tasks, \ourmodel outperforms the unrefined LLM, with the smallest difference being on the reverse facts for UMLS where \ourmodel noticeably does not learn to infer the inverse relationships from training. For UMLS fact recall \ourmodel demonstrates $100\%$ accuracy on recalling the answers to the training set and similarly performs extremely well on the multi-hop reasoning questions. We hypothesize that the even larger gap between LLM and \ourmodel on the multiple choice results compared to the difference on the open ended question results comes from \ourmodel learning to differentiate the good answers from poor ones even if it does not explicitly know the correct answer.

\textbf{Graph Aligned Language Models Significantly Improve Domain Knowledge Retrieval Tasks.} Large language models are extraordinary tools for general knowledge but can not produce answers to many domain specific questions modeled in complex networks. This is evidenced by \ourmodel outperforming LLM  across all domain level tasks, including simple fact retrieval questions. %Additionally, LLMs are biased towards topics with a large presence in their original training set. This is evident in the LLM often responding to co-authorship recommendation queries on the DBLP dataset with prominent and widely known scientists. 

\textbf{Increasing the Node Neighborhood Context During Training Improves Inference Performance.} Both the UMLS (Table~\ref{tab:umls}) and DBLP (Table~\ref{tab:dblp}) cases demonstrate that incorporating multiple edges into each training instance improves the language models recall and reasoning. This is evident as \ourmodel training evolves from single triple samples, to relations with multiple targets, and further to include additional neighborhood information such as the topic areas an author publishes in.

\textbf{Node Context Summarization Using LLM Improves Learning.} Using a LLM to rewrite or summarize statements produced from the node neighborhood encoding improves \ourmodelnospace's fact recall and multi-hop reasoning as shown on Table~\ref{tab:umls}. The LLM summarized version for the UMLS graph encoding outperforms other \ourmodel versions even if the same information is present in training. We postulate that variation in word choice,  mapping of node labels to more interpretable names helps significantly improve the learning process.

\section{Conclusions and Future Work}
We demonstrate an effective approach to integrate domain-specific knowledge graphs into large language models via fine-tuning. Empirically, this technique yields significant gains in multi-hop reasoning ability over the base LLM.  Our proposed fine-tuning method encodes graph structure and it's semantic knowledge into the LLM, by maximally leveraging the original LLM's strengths - textual understanding, commonsense knowledge and generative capabilities. 

In particular, quantitative experiments verify F1 score improvements of 18\% on fact recall and 13\% on complex inference queries requiring multi-hop reasoning on the UMLS domain for which the LLM already has some knowledge, and 142\% and 141\% respectively on DBLP's social network structure which represents novel information for the LLM. Given the importance of directionality of relationships in a graph, we also measure the improvement of recalling inverse facts by the resulting model. Overall, our experiments while preliminary in nature, confirm that integration via fine-tuning instills more reliable reasoning capacity based on graphs containing specialized entities and relationships, and it enables tighter coupling of structured symbolic knowledge with learned representations.  Evaluating the effectiveness of the partitioning and encoding schemes across a wider range of larger-scale graphs with highly uneven connectivity distributions are candidate for future work.

\bibliography{aaai24}

\end{document}